\documentclass{article}
\usepackage[nonatbib, final]{nips_style}
\usepackage[utf8]{inputenc} 
\usepackage[T1]{fontenc}    
\usepackage{hyperref}       
\usepackage{url}            
\usepackage{booktabs}       
\usepackage{amsfonts}       
\usepackage{nicefrac}       
\usepackage{microtype}      
\usepackage{tabularx}
\usepackage{float}
\usepackage{subcaption}
\usepackage{graphicx}
\usepackage{cite}
\usepackage{enumitem}
\usepackage[table,xcdraw]{xcolor}
\usepackage{adjustbox}
\usepackage{hyperref}
\usepackage{svg}
\usepackage{sectsty}
\usepackage{titlesec}
\usepackage{url}
\usepackage{float}
\usepackage{microtype}
\usepackage{graphicx} 
\usepackage{amsmath}
\usepackage[utf8]{inputenc} 
\usepackage[T1]{fontenc}    
\usepackage{hyperref}       
\usepackage{url}            
\usepackage{booktabs}       
\usepackage{amsfonts}       
\usepackage{nicefrac}       
\usepackage{microtype}      
\usepackage{xcolor}         
\usepackage{xspace}
\usepackage{amsmath}
\usepackage{amssymb}
\usepackage{booktabs}
\usepackage{makecell}
\usepackage{colortbl}
\usepackage{graphicx}   
\usepackage{CJKutf8}
\usepackage{multirow}
\usepackage{interval}
\intervalconfig{soft open fences}
\usepackage{algorithm, algpseudocode}

\usepackage{float}
\usepackage{enumitem}
\usepackage{tabularx}
\usepackage{placeins}
\usepackage{color}
\usepackage{tcolorbox}
\usepackage{verbatim}
\usepackage{colortbl} 
\usepackage{listings}
\usepackage{makecell}
\usepackage{siunitx}
\usepackage{xcolor}
\definecolor{bg}{RGB}{176,226,255}
\definecolor{bonus_green}{RGB}{0,100,0}

\usepackage{tikz}
\usepackage{graphicx}
\usepackage{amssymb}
\usepackage{xcolor, color}
\usepackage{wrapfig}
\usepackage{tcolorbox}
\usepackage{pgfplots}
\usetikzlibrary{arrows.meta,positioning,calc,shapes.geometric,decorations.text,patterns,patterns.meta,fit,backgrounds,chains,shadows,math,circuits.ee.IEC,decorations.pathmorphing, decorations.pathreplacing, decorations.shapes}

\title{Universal Reasoning Model}

\author{
\textbf{Zitian Gao} \quad
\textbf{Lynx Chen} \quad
\textbf{Yihao Xiao} \quad
\textbf{He Xing} \quad
\textbf{Ran Tao} \\ \\ \quad
\textbf{Haoming Luo} \quad
\textbf{Joey Zhou} \quad
\textbf{Bryan Dai} \thanks{~Corresponding author.} \vspace{4mm} \\
\hspace{5mm} Ubiquant \quad \vspace{3mm} \\ \small \hspace{3mm} \texttt{\{ztgao02,ylchen,yhxiao,xyyang,rtao02,hmluo,jzhou,cbdai\}}\\ \hspace{3mm} \texttt{@ubiquant.com}
}

\begin{document}
\maketitle

\begin{abstract}

Universal transformers (UTs) have been widely used for complex reasoning tasks such as ARC-AGI and Sudoku, yet the specific sources of their performance gains remain underexplored. In this work, we systematically analyze UTs variants and show that improvements on ARC-AGI primarily arise from the recurrent inductive bias and strong nonlinear components of Transformer, rather than from elaborate architectural designs. Motivated by this finding, we propose the Universal Reasoning Model (URM), which enhances the UT with short convolution and truncated backpropagation. Our approach substantially improves reasoning performance, achieving state-of-the-art$^{\ast}$ 53.8\% pass@1 on ARC-AGI 1 and 16.0\% pass@1 on ARC-AGI 2. \footnotetext{$^{\ast}$This comparison focuses on pass@1 score of single small models trained
from scratch under the same data setting as HRM and TRM, excluding test-time scaling, ensembling, and visual methods such as VARC~\cite{VARC}.} Our code is avaliable at \url{https://github.com/UbiquantAI/URM}.

\end{abstract}
\section{Introduction}
Recent advances in recurrent models \cite{hrm1, hrm2, trm} have demonstrated the effectiveness of Universal Transformers (UTs)~\cite{ut}  in addressing complex reasoning tasks, such as ARC-AGI and Sudoku \cite{arc1, arc2}. UT-based small models, despite being trained from scratch on these tasks without internet-scale pre-training, consistently outperform most standard Transformer-based Large Language models (LLMs) by a significant margin~\cite{hrm1}.

\begin{figure}[H]
    \centering
    \includegraphics[width=1\linewidth]{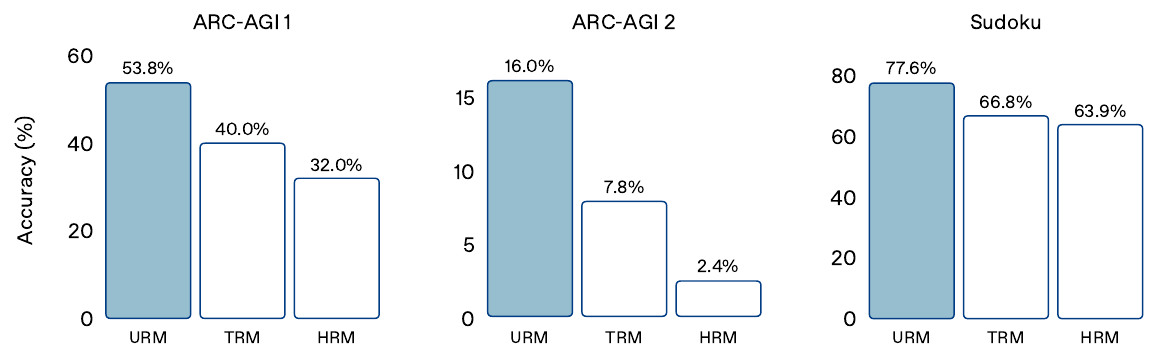}
    \caption{Performance comparison of UT-based models on the ARC-AGI and Sudoku benchmarks. ARC-AGI 1 and 2 scores are taken from the official ARC-AGI leaderboard for reliability.}
    \label{fig:fig1}
\end{figure}

While this contrast highlights the potential of UTs for depth-intensive iterative reasoning, the function and impact of gating mechanisms remain insufficiently explored beyond their initial intuition




Prior studies often attribute improvements to high-level architectural innovations~\cite{hrm1,hrm2,trm}, yet our analysis reveals that the core performance gain actually arises from the often-overlooked recurrent inductive bias intrinsic to the Universal Transformer. In particular, nonlinear depth-wise computation plays a much larger role than previously acknowledged, suggesting that architectural modifications that enhance recurrent processing can yield substantial downstream improvements. Motivated by this insight, we further investigate and strengthen this inductive bias via a simplified yet effective enhancement to the UT framework, enabling stronger abstraction capabilities while preserving parameter efficiency.

Our main contributions are as follows:
\vspace{-1mm}
\begin{itemize}[leftmargin=*]
    \item Through extensive ablation studies, we show that the performance of models on ARC-AGI–style complex reasoning tasks primarily stems from their nonlinearity. Moreover, we reveal that the true source of reasoning capability beyond standard Transformers comes from the recurrent mechanism of Universal Transformers rather than overly elaborate design in prior work.
    \item By introducing short convolutions and truncated backpropagation into the Universal Transformer, we achieve a state-of-the-art 53.8\% pass@1 accuracy on ARC-AGI 1 and 16.0\% on ARC-AGI 2.
\end{itemize}

\section{Preliminaries}

\subsection{Standard Transformer}
Let $\mathcal{V}$ denote the vocabulary of size $V$, and let $\mathbf{x} = (x_1, \dots, x_N) \in \mathcal{V}^N$ be an input sequence of length $N$. We define the token embedding function as $\phi: \mathcal{V}^N \to \mathbb{R}^{N \times d}$, mapping discrete tokens to a $d$-dimensional continuous representation. Conversely, the unembedding function (or language modeling head) is denoted by $\psi: \mathbb{R}^{N \times d} \to \mathbb{R}^{N \times V}$, which projects hidden states back to the vocabulary logit space.

A single Transformer layer, parameterized by $\theta$, is defined as a function $\mathcal{T}_\theta: \mathbb{R}^{N \times d} \to \mathbb{R}^{N \times d}$. This function typically composes a Multi-Head Self-Attention (MHSA) module and a Position-wise Feed-Forward Network (FFN), each wrapped with residual connections and layer normalization:

\begin{equation*}
    \begin{split}
        \mathcal{T}_\theta(H) &= \text{FFN}(\text{LN}(H' + H)), \\
        \text{where } H' &= \text{MHSA}(\text{LN}(H))
    \end{split}
\end{equation*}

A standard, non-recursive Transformer model $\mathcal{M}_{\text{std}}$ of depth $L$ is constructed by stacking $L$ layers with distinct parameters $\Theta = \{\theta_1, \dots, \theta_L\}$. The forward pass is the composition of these layers:

\begin{equation*}
    \mathcal{M}_{\text{std}}(\mathbf{x}) = \psi \circ \mathcal{T}_{\theta_L} \circ \dots \circ \mathcal{T}_{\theta_1} \circ \phi(\mathbf{x})
\end{equation*}

Here, the operator $\circ$ denotes function composition. The computational cost and parameter count both scale linearly with $L$, creating a rigid coupling between model capacity and inference compute.

\subsection{Universal Transformer}

The Universal Transformer (UT) \cite{ut} extends the standard Transformer \cite{attn} by introducing \emph{recurrent computation over depth}. Instead of stacking $L$ distinct layers, the UT applies a single transition block repeatedly to refine token representations. For an input sequence $\mathbf{x}$ with embedding matrix $\mathbf{H}^0 \in \mathbb{R}^{n\times d}$, the UT updates states as

\begin{equation*}
    \mathbf{H}^{t+1} = \mathrm{LayerNorm}\!\left(
        \mathbf{H}^{t} + \mathrm{MHA}\!\left(\mathbf{H}^{t}\right)
    \right),
\end{equation*}

followed by a shared position-wise transition function

\begin{equation*}
    \mathbf{H}^{t+1} \leftarrow \mathrm{LayerNorm}\!\left(
        \mathbf{H}^{t+1} + \mathrm{Transition}\!\left(\mathbf{H}^{t+1}\right)
    \right), \qquad t = 0,\dots,T-1,
\end{equation*}

where $\mathrm{Transition}$ is either a feed-forward network or separable convolution. To encode both position and refinement depth, UT adds 2-D sinusoidal embeddings at each step.

\subsubsection{Parameter Sharing}

A key design of UT is \emph{weight tying} across depth. The attention and transition parameters

\begin{equation*}
    \Theta_{\mathrm{UT}} = \{ \mathbf{W}_h^Q, \mathbf{W}_h^K, \mathbf{W}_h^V, \mathbf{W}^O,
    \Theta_{\mathrm{Transition}} \}
\end{equation*}

are reused for all $t$. Thus, the model performs iterative representation refinement with a flexible number of steps $T$, enabling (i) depth adaptation at inference and (ii) higher theoretical expressivity than fixed-depth Transformers.

\subsubsection{Adaptive Computation Time (ACT)}

With ACT \cite{act}, different tokens may halt at different recurrent steps. At step $t$, each position predicts a halting probability

\begin{equation*}
    p_{t,i} = \sigma(\mathbf{w}^\top \mathbf{h}_{t,i} + b),
\end{equation*}

accumulated until reaching threshold $1-\epsilon$. The final token representation is a weighted mixture

\begin{equation*}
    \mathbf{h}^{\mathrm{final}}_{i} = \sum_{t} \Delta_{t,i}\,\mathbf{h}_{t,i},
\end{equation*}

where $\Delta_{t,i}$ is the truncated allocation. ACT allows UT to allocate more computation to complex tokens and less to simpler ones.

\section{Universal Reasoning Model}

The base architecture of our Universal Reasoning Model (URM) closely follows that of the Universal Transformer \cite{ut}, with the difference being its decoder-only design. This aspect is consistent with previous works such as HRM \cite{hrm1} and TRM \cite{trm}. Our work differs from previous models \cite{hrm1,trm} by introducing the following \textbf{ConvSwiGLU} module and a \textbf{Truncated Backpropagation Through Loops} mechanism.

\subsection{ConvSwiGLU}
\label{convswiglu}
To strengthen the non-linearity of Universal Transformer, we introduce a \textbf{ConvSwiGLU} (motivation see Section~\ref{nonlinear}), which augments the standard SwiGLU feed-forward block with a depthwise short convolution. Unlike the conventional point-wise SwiGLU \cite{glu}, which treats each token independently, our design explicitly injects \emph{local contextual interactions} into the gating mechanism, introducing lightweight channel mixing in token space without increasing sequence-level complexity \cite{plm,metaformer}. 

Given an input sequence $X \in \mathbb{R}^{T \times d}$, we first project it into an expanded intermediate representation:

\begin{equation*}
[\mathbf{G}, \mathbf{U}] = X W_{\text{up}} \in \mathbb{R}^{T \times 2m}.
\end{equation*}

The SwiGLU activation produces a gated representation:

\begin{equation*}
\mathbf{H}_{\text{ffn}} = \text{SiLU}(\mathbf{G}) \odot \mathbf{U}.
\end{equation*}

To integrate short-range token interactions, we apply a \emph{depthwise 1D convolution} over the gated features:

\begin{equation*}
\mathbf{H}_{\text{conv}} 
= \sigma \bigl( \mathbf{W}_{\text{dwconv}} * \mathbf{H}_{\text{ffn}} \bigr),
\end{equation*}

where $\mathbf{W}_{\text{dwconv}} \in \mathbb{R}^{m \times 1 \times k}$ is a depthwise convolution kernel of size $k=2$.

Finally, the output is projected back to the hidden dimension:

\begin{equation*}
\boxed{
\mathbf{Y}
= \bigl[ \sigma ( \mathbf{W}_{\text{dwconv}} * (\text{SiLU}(\mathbf{G}) \odot \mathbf{U}) ) \bigr]
W_{\text{down}}.
}
\end{equation*}

\usetikzlibrary{arrows.meta,positioning,calc,shapes.geometric}
\definecolor{delta_color}{RGB}{242,243,193}
\definecolor{swa_color}{RGB}{252,224,225}
\definecolor{glu_color}{RGB}{194,232,247}
\definecolor{silu_color}{RGB}{203,231,207}
\definecolor{linear_color}{RGB}{220,223,240}
\definecolor{conv_color}{RGB}{252,224,225}
\definecolor{gray_bbox_color}{RGB}{243,243,244}
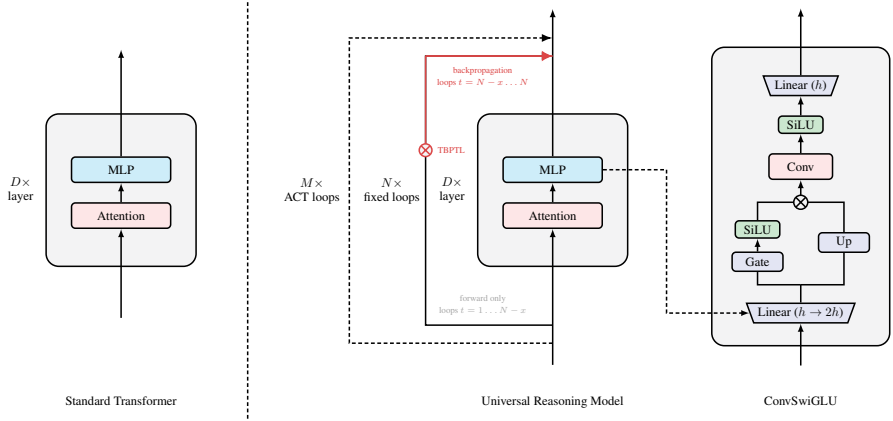
\begin{figure}[t!]
\centering
\scalebox{0.85}{
\tikzset{
  model/.style={
    draw=black, very thick, fill=gray_bbox_color,
    minimum width=118pt, rounded corners=10pt
  },
  gdelta/.style={
    draw=black, very thick, fill=gray_bbox_color,
    minimum width=120pt, minimum height=190pt, rounded corners=10pt
  },
  swa/.style={
    draw=black, very thick, fill=swa_color!80,
    minimum width=78pt, minimum height=0.7cm, rounded corners=3pt
  },
  glu/.style={
    draw=black, very thick, fill=glu_color!80,
    minimum width=78pt, minimum height=0.7cm, rounded corners=3pt
  },
  conv/.style={
    draw=black, very thick, minimum width=50pt, minimum height=20pt,
    fill=conv_color!80, rounded corners=3pt
  },
  linear/.style={
    draw=black, very thick, trapezium,
    trapezium left angle=110, trapezium right angle=110,
    minimum width=50pt, minimum height=15pt, fill=linear_color!80
  },
  branch_node/.style={
    draw=black, very thick, minimum width=40pt, minimum height=15pt,
    fill=linear_color!80, rounded corners=3pt
  },
  silu/.style={
    draw=black, very thick, fill=silu_color,
    minimum width=35pt, minimum height=12pt, rounded corners=3pt
  },
  layerlink/.style={-latex, very thick},
  modulelink/.style={-latex, very thick, densely dashed},
  normlink/.style={very thick},
  otimes/.style={
    draw=black, very thick, circle, minimum size=11pt,
    inner sep=0pt, outer sep=0pt,
    path picture={
      \draw (path picture bounding box.center) -- ++(0.25cm,0.25cm)
            (path picture bounding box.center) -- ++(-0.25cm,-0.25cm)
            (path picture bounding box.center) -- ++(-0.25cm,0.25cm)
            (path picture bounding box.center) -- ++(0.25cm,-0.25cm);
    }
  },
  gradlink/.style={very thick, grad_color, -{Latex[length=3mm]}},
  stopgrad/.style={
    fill=gray_bbox_color,
    draw=grad_color, very thick, circle, minimum size=10pt, inner sep=0pt,
    path picture={
      \draw[grad_color, very thick]
        (path picture bounding box.south west) -- (path picture bounding box.north east)
        (path picture bounding box.north west) -- (path picture bounding box.south east);
    }
  }
}

\definecolor{grad_color}{RGB}{220,70,70}

\centering
\hspace*{-25pt}
\resizebox{\textwidth}{!}{
\begin{tikzpicture}

\node[model, minimum height=120pt] (model1) at (0,0) {};
\node[left=6pt, align=center] at (model1.west) {$D\times$ \\ layer};
\coordinate (left_input_start) at ($(model1.south) + (0, -40pt)$);
\node[below=58pt] at (left_input_start) {Standard Transformer};
\node[swa, anchor=south, yshift=30pt] at (model1.south) (attn) {Attention};
\draw[layerlink] (left_input_start) -- (attn.south);
\node[glu, anchor=south, yshift=15pt] at (attn.north) (urm) {MLP};
\draw[layerlink] (attn.north) -- (urm.south);
\coordinate (left_output_end) at ($(model1.north) + (0, 50pt)$);
\draw[layerlink] (urm.north) -- (left_output_end);

\draw[densely dashed, very thick] ($(model1.east)+(40pt, -180pt)$) -- ($(model1.east)+(40pt, 150pt)$);

\node[model, minimum height=120pt] (model2) at ($(model1.east)+(280pt,0pt)$) {};
\node[left=6pt, align=center] at (model2.west) {$D\times$ \\ layer};
\coordinate (right_input_start) at ($(model2.south) + (0, -77pt)$);
\node[below=21pt] at (right_input_start) {Universal Reasoning Model};
\node[swa, anchor=south, yshift=30pt] at (model2.south) (attn2) {Attention};
\draw[layerlink] (right_input_start) -- (attn2.south);
\node[glu, anchor=south, yshift=15pt] at (attn2.north) (urm2) {MLP};
\draw[layerlink] (attn2.north) -- (urm2.south);
\coordinate (right_output_end) at ($(model2.north) + (0, 82pt)$);
\draw[layerlink] (urm2.north) -- (right_output_end);

\coordinate (loop_exit2)  at ($(right_input_start) + (-0pt, 1.1)$);
\coordinate (loop_left2)  at ($(loop_exit2) + (-100pt, 0)$);
\coordinate (loop_up2)    at ($(right_output_end) + (-100pt, -1.3)$);
\coordinate (loop_entry2) at ($(right_output_end) + (-0pt, -1.3)$);
\draw[very thick, -{Latex[length=3mm]}]
  (loop_exit2) -- (loop_left2) -- (loop_up2) -- (loop_entry2);
\node[left=1pt, align=center, yshift=0pt] 
    at ($(loop_left2)!0.5!(loop_up2)$) {$N\times$ \\ fixed loops};

\coordinate (tbptl_cut_fixed) at ($(loop_left2)!0.65!(loop_up2)$);

\draw[very thick, grad_color] (loop_entry2) -- (loop_up2);

\draw[very thick, grad_color] (loop_up2) -- (tbptl_cut_fixed);

\draw[very thick, grad_color, -{Latex[length=3mm,width=3mm]}]
  (loop_up2) -- (loop_entry2);

\node[stopgrad] at (tbptl_cut_fixed) {};

\node[grad_color, anchor=west] at ($(tbptl_cut_fixed)+(6pt,0pt)$) {\scriptsize TBPTL};
\node[grad_color, align=center]
  at ($(tbptl_cut_fixed)!0.55!(loop_entry2)+(-10pt,17pt)$)
  {\scriptsize backpropagation\\[-1pt]\scriptsize loops $t=N-x\ldots N$};

\node[gray!70, align=center]
  at ($(loop_exit2)!0.55!(tbptl_cut_fixed)+(0pt,-60pt)$)
  {\scriptsize forward only\\[-1pt]\scriptsize loops $t=1\ldots N-x$};

\coordinate (loop_exit_outer2) at ($(loop_exit2) + (-0pt, -0.5)$);
\coordinate (loop_left_outer2) at ($(loop_left2) + (-60pt, -0.5)$);
\coordinate (loop_up_outer2)   at ($(loop_up2) + (-60pt, 0.5)$);
\coordinate (loop_entry_outer2) at ($(loop_entry2) + (-0pt, 0.5)$);
\draw[densely dashed, very thick, -{Latex[length=3mm]}]
    (loop_exit_outer2) -- (loop_left_outer2) -- (loop_up_outer2) -- (loop_entry_outer2);
\node[left=4pt, align=center, yshift=0pt]
    at ($(loop_left_outer2)!0.5!(loop_up_outer2)$) {$M\times$ \\ ACT loops};

\node[gdelta,
      anchor=west,
      xshift=65pt,
      yshift=-5pt,
      minimum width=140pt,
      minimum height=235pt]
  (convswiglu_block) at (model2.east) {};

\coordinate (right_input_start2) at ($(convswiglu_block.south) + (0, -15pt)$);
\node[below=21pt] at (right_input_start2) {ConvSwiGLU};

\node[linear, anchor=south, yshift=18pt]
  at (convswiglu_block.south) (gate_up)
  {Linear ($h \to 2h$)};
\draw[layerlink] (right_input_start2) -- (gate_up.south);

\coordinate (split_point) at ($(gate_up.north) + (0, 14pt)$);
\draw[normlink] (gate_up.north) -- (split_point);

\node[branch_node,
      anchor=south,
      xshift=-34pt,
      yshift=10pt]
  at (split_point) (gate_chunk) {Gate};
\draw[normlink] (split_point) -| (gate_chunk.south);

\node[branch_node,
      anchor=south,
      xshift=34pt,
      yshift=25pt]
  at (split_point) (up_chunk) {Up};
\draw[normlink] (split_point) -| (up_chunk.south);

\node[silu,
      anchor=south,
      yshift=10pt]
  at (gate_chunk.north) (silu_gate) {SiLU};
\draw[layerlink] (gate_chunk.north) -- (silu_gate.south);

\coordinate (merge_point_y) at ($(silu_gate.north) + (0, 14pt)$);
\node[otimes]
  at (gate_up.north |- merge_point_y) (multiply) {};
\draw[normlink] (silu_gate.north) |- (multiply.west);
\draw[normlink] (up_chunk.north)  |- (multiply.east);

\node[conv,
      anchor=south,
      yshift=12pt]
  at (multiply.north) (conv)
  {Conv};
\draw[layerlink] (multiply.north) -- (conv.south);

\node[silu,
      anchor=south,
      yshift=15pt]
  at (conv.north) (silu_act) {SiLU};
\draw[layerlink] (conv.north) -- (silu_act.south);

\node[linear,
      anchor=south,
      yshift=15pt]
  at (silu_act.north) (down_proj)
  {Linear ($h$)};
\draw[layerlink] (silu_act.north) -- (down_proj.south);

\coordinate (right_output_end2)
  at ($(convswiglu_block.north) + (0, 30pt)$);
\draw[layerlink] (down_proj.north) -- (right_output_end2);

\draw[modulelink]
  (urm2.east) -- ++(50pt,0) |- (gate_up.west);

\end{tikzpicture}
}
}
\vspace{5mm}
\caption{Illustration of our Universal Reasoning Model (URM) architecture. The left shows a standard Transformer layer stack, while the right illustrates the URM with fixed loops, ACT loops, and the ConvSwiGLU module. For illustrative purposes, components such as embeddings, residual connections, RMSNorm, positional encodings, and other modules are omitted, x in right figure represents the first x loops of the inner loop in forward-only mode, TBPTL represents our proposed Truncated Backpropagation Through Loops.}
\label{fig:urm_left}
\end{figure}

\subsection{Truncated Backpropagation Through Loops}
\label{sec:truncated-bptt}

When the number of recurrent reasoning loops becomes large, the gradients propagated from early loops may hinder optimization due to noise accumulation and instability (see empirical evidence in Section~\ref{tbp}). To alleviate this issue, we employ \textbf{Truncated Backpropagation Through Loops (TBPTL)} and only compute gradients for the later loops.

\vspace{0.5em}
\noindent
Consider a $D$-layer Universal Reasoning Model unrolled for $M$ iterative loops during training. Let $\mathbf{h}_{t}^{(d)}$ denote the hidden representation of layer $d \in \{1,\ldots,D\}$ at iteration $t \in \{1,\ldots,M\}$. The recurrent transition is defined as:

\begin{equation*}
    \mathbf{h}_{t}^{(d)} = F_{\theta}^{(d)}\big(\mathbf{h}_{t}^{(d-1)}, \mathbf{h}_{t-1}^{(d)}\big),
    \label{eq:recur_update}
\end{equation*}

where $F_{\theta}^{(d)}$ denotes the parameterized transformation at layer $d$ with trainable parameters $\theta$.

\vspace{0.5em}
\noindent
Instead of backpropagating through all $M$ loops, we partition the rollout into \emph{forward-only} and \emph{trainable} segments. Specifically, for a truncation index $N < M$:
\begin{equation*}
    \underbrace{\{1,2,\ldots,N\}}_{\text{no backward pass}}, 
    \qquad
    \underbrace{\{N+1,\ldots,M\}}_{\text{forward + backward}}.
\end{equation*}
During training, we compute gradients only on the loss accumulated in the latter $(M-N)$ loops:
\begin{equation*}
    \mathcal{L}_{\text{TBPTL}}(\theta) 
    = \sum_{t=N+1}^{M} \mathcal{L}\big(\mathbf{h}_{t}^{(D)}, y\big),
    \label{eq:tbptt_loss}
\end{equation*}
where $\mathcal{L}(\cdot)$ is cross-entropy loss function. The gradients with respect to $\theta$ are thus:
\begin{equation*}
    \nabla_{\theta} \mathcal{L}_{\text{TBPTL}}
    = \sum_{t=N+1}^{M} 
    \frac{\partial \mathcal{L}}{\partial \mathbf{h}_{t}^{(D)}}
    \frac{\partial \mathbf{h}_{t}^{(D)}}{\partial \theta}.
\end{equation*}
\noindent

\section{Experiment}
\subsection{Experiment Settings}
\label{settings}
Our experimental setup largely follows HRM and TRM~\cite{hrm1,trm}. We use the same datasets and augmented data as in prior work, and apply an exponential moving average (EMA) to model parameters to improve training stability, following~\cite{trm}. All models are trained with the AdamAtan2 optimizer~\cite{adamatan2}. For ARC-AGI 1 and ARC-AGI 2, the main model learning rates are set to $1\times10^{-4}$ and $3\times10^{-4}$, respectively, while the puzzle embedding uses a learning rate of $1\times10^{-2}$; for Sudoku, the puzzle embedding learning rate is $1\times10^{-4}$. Weight decay is set to 0.1 for both the main model and puzzle embedding on ARC-AGI 1 and ARC-AGI 2, and to 1.0 for Sudoku, consistent with prior work. The model has 4 layers with hidden size 512 and 8 attention heads. The inner loop runs for 8 steps, with the first two steps being forward-only, while the outer loop employs Adaptive Computation Time (ACT)~\cite{act} with a maximum of 16 steps.

\subsection{Main Results}

\renewcommand{\arraystretch}{1.6}
\begin{table}[H]
\centering
\resizebox{\textwidth}{!}{
\large
\begin{tabular}{c|cccc|cccc|c}
\hline
 & \multicolumn{4}{c|}{\textbf{ARC-AGI 1}} & \multicolumn{4}{c|}{\textbf{ARC-AGI 2}} & \textbf{Sudoku} \\
 & pass@1 & pass@10 & pass@100 & pass@1000 & pass@1 & pass@10 & pass@100 & pass@1000 & pass@1 \\
\hline
\textbf{HRM} & 34.4 & 46.4 & 55.0 & 60.5 & 5.4 & 9.6 & 14.3 & 18.6 & 63.9 \\
\textbf{TRM} & 40.0 & 51.3 & 59.8 & 64.4 & 4.6 & 7.4 & 11.7 & 13.6 & 66.8 \\
\hline
\textbf{URM} & \textbf{53.8} & \textbf{71.3} & \textbf{80.4} & \textbf{85.1} & \textbf{16.0} & \textbf{26.9} & \textbf{34.3} & \textbf{41.3} & \textbf{77.6} \\
w/o Short Conv. & 45.3 & 62.6 & 72.0 & 78.3 & - & - & - & - & - \\
w/o Trunc. Backprop. & 40.0 & 54.4 & 64.5 & 70.5 & - & - & - & - & - \\
\hline
\end{tabular}
}

\vspace{5mm}
\caption{The performance of URM, TRM, and HRM on three complex reasoning tasks: ARC-AGI 1, ARC-AGI 2, and Sudoku. pass@n denotes the pass rate when sampling n answers from the model; a sample is considered correct if at least one of the n answers is correct. The scores of TRM and HRM in this table may differ from those shown in the teaser. This is because the teaser scores are taken directly from the official ARC-AGI leaderboard for rigor, whereas the scores in this table are reproduced from the official TRM and HRM repositories following their official evaluation procedures. Minor discrepancies may occur due to randomness.}
\end{table}

As shown in Table 1, the Universal Reasoning Model (URM) achieves substantial improvements over prior UT-based approaches across all benchmarks. On ARC-AGI 1, URM reaches 53.8\% pass@1, outperforming TRM (40.0\%) and HRM (34.4\%) by large margins. On ARC-AGI 2, URM obtains 16.0\% pass@1, nearly tripling HRM and more than doubling TRM. A similar advantage appears on Sudoku, where URM achieves 77.6\% accuracy, surpassing both TRM and HRM.

Notably, URM’s gains further widen under larger sampling budgets (e.g., pass@1000), indicating that iterative refinement enables richer candidate generation rather than brittle one-step predictions.

\subsection{Why Universal Transformer?}

\begin{table}[H]
\centering

\small
\renewcommand{\arraystretch}{1.2}

\begin{tabular*}{\textwidth}
{@{\hskip\tabcolsep\extracolsep{\fill}} c c c c c c c c c}
\hline
\textbf{Layer} & \textbf{Loop} & \textbf{Hidden Size} & \textbf{Params} & \textbf{FLOPs} &
\textbf{pass@1} & \textbf{pass@10} & \textbf{pass@100} & \textbf{pass@1000} \\
\hline
\midrule
\multicolumn{9}{c}{Vanilla Transformers} \\
\midrule
2  & 1 & 256  & $1\times$   & $1\times$   & 0.75  & 3.75  & 5.75  & 7.00  \\
2  & 1 & 384  & $1.5\times$ & $1.5\times$ & 2.75  & 4.13  & 6.75  & 9.13  \\
2  & 1 & 512  & $2\times$   & $2\times$   & 3.63  & 6.00  & 8.88  & 11.00 \\
2  & 1 & 768  & $3\times$   & $3\times$   & 2.75  & 5.00  & 7.38  & 9.13  \\
\hline
4  & 1 & 256  & $2\times$   & $2\times$   & 4.25  & 8.25  & 10.50 & 13.88 \\
4  & 1 & 384  & $3\times$   & $3\times$   & 2.88  & 5.88  & 8.38  & 10.13 \\
4  & 1 & 512  & $4\times$   & $4\times$   & 5.13  & 9.00  & 10.50 & 12.63 \\
4  & 1 & 768  & $6\times$   & $6\times$   & 5.63  & 9.25  & 10.75 & 12.25 \\
\hline
6  & 1 & 256  & $3\times$   & $3\times$   & 4.63  & 8.75  & 11.75 & 13.38 \\
6  & 1 & 384  & $4.5\times$ & $4.5\times$ & 5.00  & 9.38  & 11.25 & 13.25 \\
6  & 1 & 512  & $6\times$   & $6\times$   & 7.88  & 11.13 & 13.75 & 15.63 \\
6  & 1 & 768  & $9\times$   & $9\times$   & 8.13  & 12.13 & 16.13 & 17.88 \\
\hline
8  & 1 & 256  & $4\times$   & $4\times$   & 6.88  & 11.25 & 13.63 & 15.63 \\
8  & 1 & 384  & $6\times$   & $6\times$   & 7.00  & 11.38 & 13.13 & 14.63 \\
8  & 1 & 512  & $8\times$   & $8\times$   & 8.50  & 12.75 & 15.75 & 17.13 \\
8  & 1 & 768  & $12\times$  & $12\times$  & 10.63 & 17.38 & 21.50 & 23.25 \\
\hline
16 & 1 & 1024 & $32\times$  & $32\times$  & 0.00  & 6.50  & 8.75  & 9.75  \\
32 & 1 & 512  & $32\times$  & $32\times$  & 23.75 & 34.13 & 38.88 & 43.38 \\
64 & 1 & 256  & $32\times$  & $32\times$  & 18.25 & 31.75 & 38.25 & 41.38 \\
\hline
\midrule
\multicolumn{9}{c}{Universal Transformers} \\
\midrule
2 & 8 & 512 & $2\times$ & $16\times$ & 36.25 & 50.75 & 61.25 & 66.88 \\
4 & 8 & 512 & $4\times$ & $32\times$ &
\textbf{40.00} & \textbf{54.38} & \textbf{64.50} & \textbf{70.50} \\
\hline

\end{tabular*}

\vspace{5mm}
\caption{Comparison between vanilla Transformers and Universal Transformers under different model depths, hidden sizes, and loops. We report pass@$n$ results on ARC-AGI 1.}
\label{why}
\end{table}

Table~\ref{why} demonstrates that the performance gains of Universal Transformers (UTs) on ARC-AGI 1 arise from substantially higher parameter efficiency rather than increased model scale or computational budget. With only 4× parameters, a UT achieves a pass@1 score of 40.0, dramatically outperforming vanilla Transformers that employ up to 32× more parameters yet remain markedly weaker. Simply scaling depth or width in vanilla Transformers yields diminishing returns and can even lead to performance degradation, highlighting a fundamental inefficiency in how parameters are used to support multi-step reasoning.

Crucially, this advantage persists even when computation is held constant. At 32× FLOPs, reallocating computation from deep, non-shared layers to recurrent refinement improves pass@1 from 23.75 for vanilla Transformers to 40.0 for UTs. This behavior is consistent with analyses of previous works~\cite{onthepower}, which argue that many reasoning tasks benefit more from iterative computation than from increasing the number of independent layers. In standard Transformers, additional FLOPs are often spent on redundant refinement in higher layers, whereas recurrent computation converts the same budget into increased effective depth~\cite{importance,onthepower}.

This superior efficiency is driven by the recurrent inductive bias introduced by parameter sharing across depth. Through repeated application of a shared transformation, UTs realize iterative refinement that better aligns with the structure of algorithmic reasoning, while avoiding any increase in parameter count. Consequently, under both fixed parameter and fixed FLOPs budgets, UTs consistently outperform vanilla Transformers on reasoning tasks, making them particularly well suited for reasoning-intensive settings such as ARC-AGI, where multi-step abstraction is more critical than sheer scale.

\subsection{Short Convolution}
\label{conv}

\begin{figure}[H]
    \centering
    \includegraphics[width=1.0\textwidth]{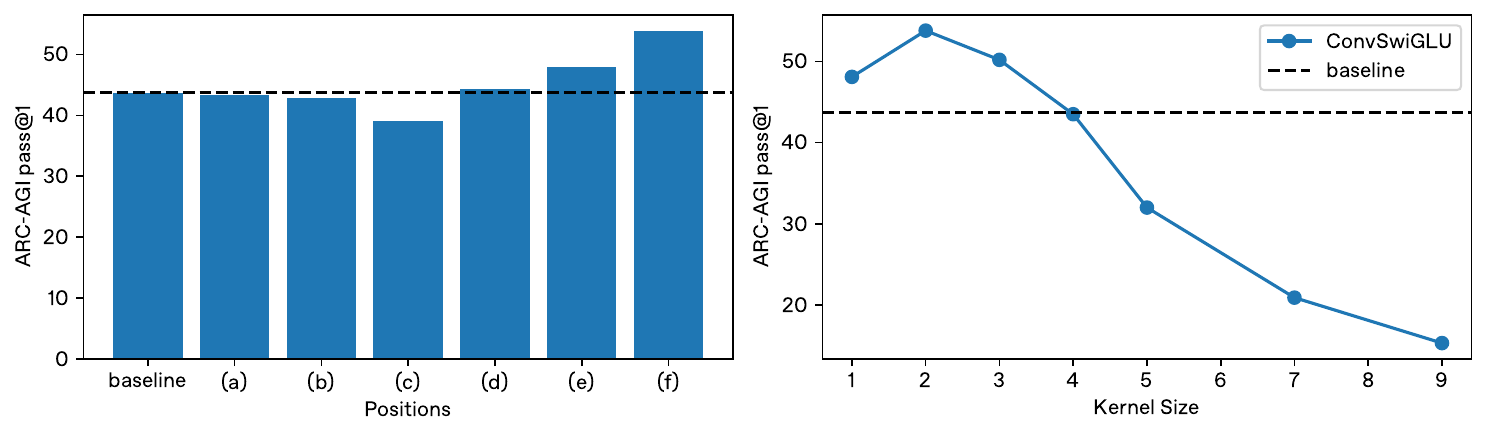}
    \caption{ARC-AGI pass@1 results for inserting the short convolution module at different positions within the UT transition (left figure), and varying the kernel size of the ConvSwiGLU module applied after the MLP expansion (right figure).}
    \label{fig:shortconv}
\end{figure}

To strengthen the nonlinear inductive bias of the Universal Transformer, we introduce a 
depthwise short convolution module parameterized by $W_{\mathrm{dwconv}}$ (see Section~\ref{convswiglu} for details), which provides 
token-local mixing while preserving the per-step computational budget. Since ARC-AGI 
performance correlates strongly with nonlinear capacity (Section~\ref{nonlinear}), we 
evaluate how inserting this module at different locations affects the recurrent transition.

We examine six insertion points: (a) after the SDPA output; (b) after the value projection; (c) after the key projection; (d) after the query projection; (e) between multi-head concatenation and the output projection; and (f) after the MLP expansion.

\begin{figure}[htbp]
    \centering
    \begin{tabular}{@{}cc@{}}
        \hspace*{-0.8cm}
        \includegraphics[width=0.53\linewidth]{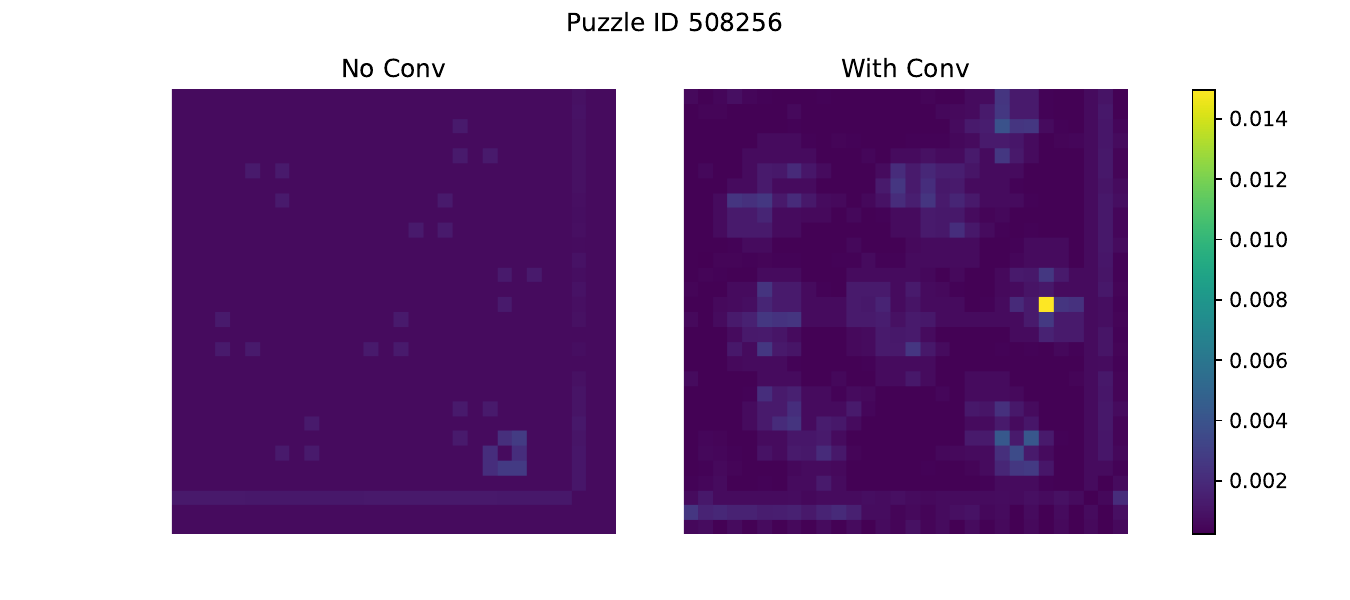} &
        
        \hspace*{-0.2cm} \includegraphics[width=0.53\linewidth]{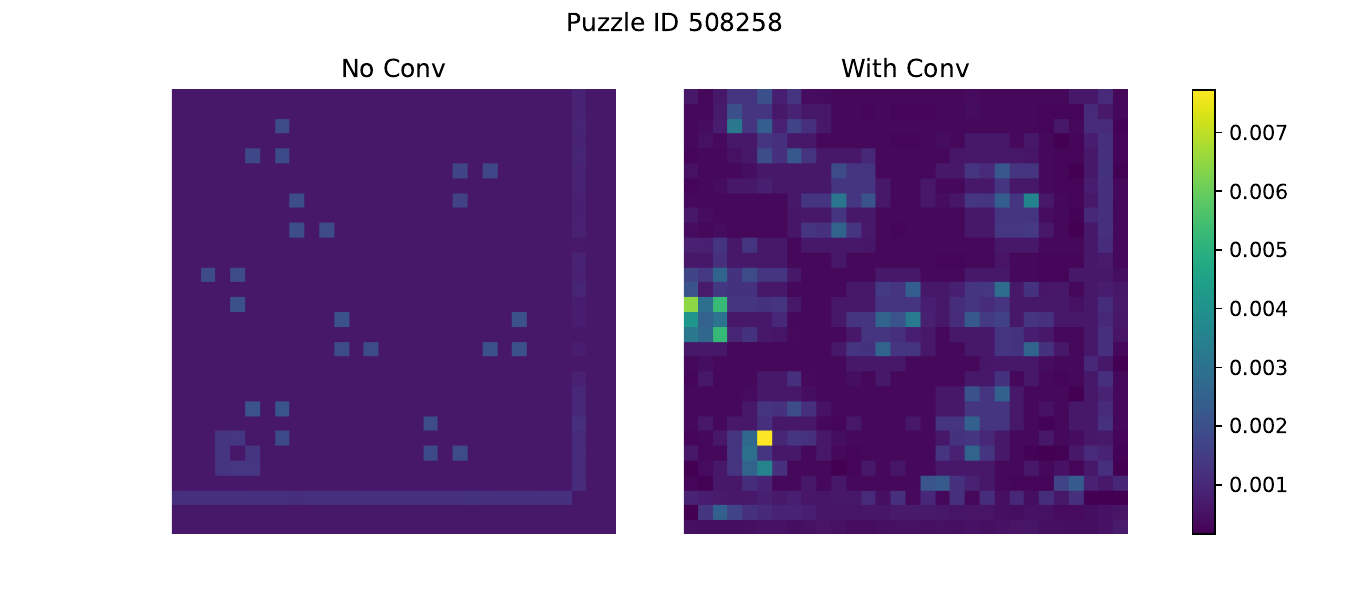}
    \end{tabular}
    \caption{Visualization of the attention matrices after adding Short Convolution. The left figure shows the standard Universal Transformer, while the right figure shows the Universal Transformer with ConvSwiGLU applied.}
    \label{fig:channel}
\end{figure}

As shown in Figure~\ref{fig:shortconv}, inserting the $W_{\mathrm{dwconv}}$ module 
inside the attention pathway, positions (a)--(d), does not yield improvements and often 
degrades performance, suggesting that local perturbations interfere with the geometric 
structure of attention’s linear projections. A mild gain appears at position (e), where the 
perturbation acts only on aggregated multi-head features.

The dominant effect arises at position (f), after the MLP expansion, indicating that short-range 
mixing is most beneficial when applied within an already nonlinear subspace. This supports a 
functional interpretation in which the MLP—not attention—constitutes the model’s primary 
source of expressive nonlinearity; augmenting it with $W_{\mathrm{dwconv}}$ substantially 
enhances the model’s nonlinear representational capacity.

As shown in Fig.~\ref{fig:channel}, the incorporation of short convolution into the MLP significantly enhances channel mixing. While the standard Universal Transformer exhibits relatively sparse and homogeneous attention patterns, the model with ConvSwiGLU produces attention matrices with more diverse and structured distributions. This suggests that short convolution facilitates more effective inter-channel information flow, thereby improving the expressiveness of the attention mechanism.

\subsection{Truncated Backpropagation Through Loops}
\label{tbp}

\renewcommand{\arraystretch}{1.4}
\begin{table}[H]
\centering
\small
\begin{tabular}{c c c c c c}
\hline
\textbf{Loop w/ grad.} & \textbf{Loop w/o grad.} & 
\textbf{pass@1} & \textbf{pass@10} & \textbf{pass@100} & \textbf{pass@1000} \\
\hline
8 & 0 & 36.25 & 50.75 & 61.25 & 66.88 \\
7 & 1 & 37.75 & 49.13 & 59.50 & 65.88 \\
6 & 2 & 39.13 & \textbf{53.50} & \textbf{61.88} & \textbf{66.88} \\
5 & 3 & \textbf{39.50} & 51.63 & 60.88 & 65.25 \\
4 & 4 & 38.75 & 50.50 & 61.50 & 65.88 \\
3 & 5 & 36.88 & 49.00 & 57.75 & 63.88 \\
2 & 6 & 34.25 & 46.25 & 55.75 & 61.75 \\
1 & 7 & 22.50 & 37.00 & 45.38 & 52.38 \\
\hline
\end{tabular}

\vspace{3mm}
\caption{Effect of Truncated Backpropagation Through Loops (TBPTL) across inner loops on ARC-AGI 1. ``Loop w/o grad.'' denotes the number of forward-only inner-loop iterations, while ``Loop w/ grad.'' indicates the number of inner loops involved in backpropagation.}
\label{tab:tbp_loops}
\end{table}

\begin{figure}[H]
    \centering
    \includegraphics[width=1.0\textwidth]{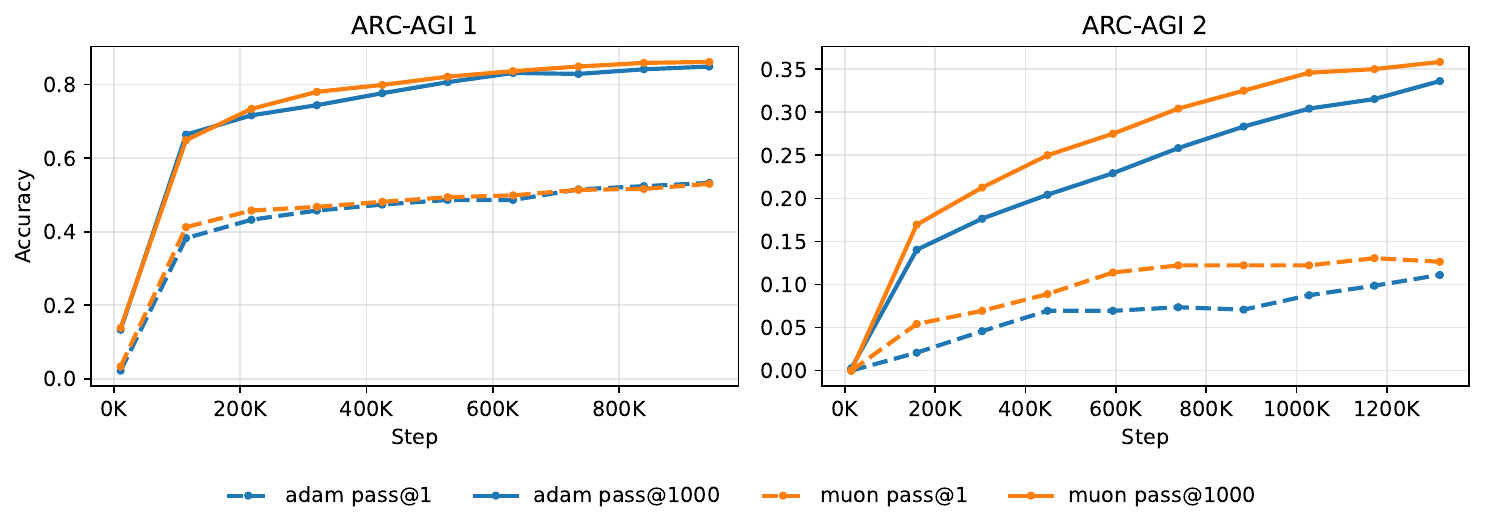}
    \caption{ARC-AGI pass@1 and pass@1000 performance of Adam and Muon optimizers on ARC-AGI 1 and ARC-AGI 2 benchmarks. Solid lines denote pass@1000, dashed lines denote pass@1, and colors indicate different optimizers. Training steps are shown in thousands (K).}
    \label{fig:tbp}
\end{figure}

As shown in Table~\ref{fig:tbp}, when the total number of inner loops is fixed to 8, truncating gradients for the first two loops—i.e., running the initial two inner-loop iterations in forward-only mode—achieves the best performance. Both pass@1 and pass@1000 peak at this truncation setting, while shorter or longer truncation horizons result in inferior outcomes.

This trend closely resembles truncated backpropagation through time (TBPTT) in recurrent neural networks, where the underlying motivation is largely the same. In full backpropagation through time, gradients are propagated through the entire sequence, which incurs high computational and memory costs and often yields ineffective long-range gradients due to vanishing or exploding behaviors. As a result, practical implementations typically restrict gradient propagation to a fixed recent window, e.g., by backpropagating errors only through the last $L$ time steps and updating the network parameters accordingly~\cite{difficulty,tbptt}.

Similarly, in universal transformers, propagating gradients across all inner-loop iterations can lead to unstable optimization, while overly aggressive truncation limits the model’s ability to coordinate multi-step refinement. Moderately truncating gradient propagation therefore provides a favorable balance between optimization stability and effective long-horizon learning.

We note that all results in this experiment are obtained using a two-layer URM without the short convolution module, which differs from the full URM model reported earlier.

\subsection{Nonlinearity of Transformers}
\label{nonlinear}
\renewcommand{\arraystretch}{1.6}
\begin{table}[H]
\centering
\begin{tabular}{lcccc}
\hline
\textbf{Model} & \textbf{pass@1} & \textbf{pass@10} & \textbf{pass@100} & \textbf{pass@1000} \\
\hline
\textbf{Full Universal Reasoning Model} 
& \textbf{53.75} & \textbf{71.25} & \textbf{80.38} & \textbf{85.13} \\

w/o Short Conv. 
& 45.25 & 62.63 & 72.00 & 78.25 \\

SwiGLU $\rightarrow$ SiLU 
& 29.75 & 42.13 & 50.00 & 54.50 \\

SiLU $\rightarrow$ ReLU 
& 28.63 & 43.38 & 50.63 & 54.88 \\

w/o Attention Softmax 
& 2.00 & 6.75 & 10.25 & 15.00 \\
\hline
\end{tabular}

\vspace{5mm}
\caption{Ablation study on nonlinearity architectural components of the Universal Reasoning Model. We report pass@$n$ results on ARC-AGI 1. All experiments are conducted under exactly the same settings as in Section~\ref{settings}.}
\label{tab:ablation}
\end{table}

As shown in Table~\ref{tab:ablation}, the performance on ARC-AGI 1 decreases monotonically as nonlinear components are progressively removed from the model. Among these components, the activation function in the MLP plays a particularly critical role: replacing SwiGLU with simpler nonlinearities such as SiLU or ReLU leads to substantial degradation, while completely removing the attention softmax results in a dramatic collapse in performance. This clear monotonic trend highlights the importance of strong nonlinear transformations for solving complex abstract reasoning tasks.

These results suggest that the expressive power required for ARC-AGI primarily arises from rich nonlinear mappings. Weakening the nonlinearity may systematically limits the model’s ability to represent complex reasoning skills.

We note that the model still retains certain forms of nonlinearity that are not ablated in this study, such as the RMSNorm applied after each layer and the dot-product interaction between queries and keys in attention. However, these components are either difficult to remove without causing training instability or represent relatively weak nonlinear effects compared to explicit activation functions. As ablating them typically leads to training failure, they fall outside the scope of the present analysis.

\subsection{Muon Optimizer}
\label{sec:muon_optimizer}

To evaluate the training efficiency of the Universal Reasoning Model (URM), we compare the Muon (Momentum Updated Orthogonal Newton) optimizer\cite{jordan2024muon} with a standard adaptive baseline, Adamatan2\cite{adamatan2}. Muon approximates second-order curvature to apply orthogonal updates to better handle the complex loss landscapes\cite{gong2025makesloopedtransformersperform} induced by deep recurrent structures. Both models are trained from scratch under identical experimental settings, including batch size, learning rate schedules, and data augmentation, ensuring that any observed differences arise solely from the choice of optimizer.

Across the ARC-AGI 1 and ARC-AGI 2 benchmarks, Muon demonstrates substantially faster convergence. On ARC-AGI 2, the Muon-optimized model reaches a pass@1 accuracy of 11.5\% in approximately 600,000 training steps, whereas the Adamatan2 baseline requires over 1,300,000 steps to achieve the same performance, corresponding to nearly a twofold speedup in optimization. Despite this advantage in early training, both methods converge to similar final accuracies (approximately 53.8\% on ARC-AGI 1 and 16.0\% on ARC-AGI 2), indicating comparable asymptotic performance.

These results suggest a separation between optimization efficiency and architectural capacity in the URM. While Muon preconditions the challenging spectral properties of recurrent weight matrices\cite{liu2025muonscalablellmtraining} and reduces training cost, it does not lead to improved final generalization.
\section{Related Work}
\subsection{ARC-AGI}
Prior work on the ARC-AGI benchmark~\cite{arc1,arc2} spans vision-based formulations, large language model (LLM) adaptation, and recurrent reasoning architectures. Vision-centric approaches such as Vision ARC~\cite{VARC} reformulate ARC as an image-to-image transformation problem and show that standard visual inductive biases can achieve competitive performance, particularly with ensembling and test-time scaling. LLM-based methods explore fine-tuning and test-time training, demonstrating that transient parameter updates outperform static in-context learning on ARC-like tasks. Beyond language and vision models, recurrent architectures emphasize iterative computation as a core mechanism for abstraction. The Hierarchical Reasoning Model (HRM)~\cite{hrm1,hrm2} introduces multi-timescale recurrence and achieves strong ARC-AGI results, while subsequent analyses suggest that its gains may largely stem from recurrence rather than explicit hierarchy. The Tiny Recursive Model (TRM)~\cite{trm} further simplifies this paradigm, showing that a single lightweight network applied recursively can match or exceed more complex hierarchical designs.

\subsection{Universal Transformers (Looped Transformers)}
The Universal Transformer (UT), also known as the Looped Transformer, was introduced by Dehghani et al.~\cite{ut} as an extension of the standard Transformer with recurrent computation and adaptive computation time. Subsequent work has shown that UTs exhibit significantly stronger multi-step reasoning abilities than vanilla Transformers, as the recurrent refinement mechanism helps overcome architectural limitations in multi-hop reasoning tasks~\cite{devil,grok}. In addition, UTs demonstrate improved algorithmic learning capabilities, enabling more effective modeling of iterative and rule-based computations~\cite{algo}. By reusing parameters across refinement steps, UTs also achieve higher parameter efficiency, allowing more expressive computation without increasing model size~\cite{onthepower}.



\section{Conclusion}
We systematically investigate the sources of performance gains in Universal Transformer models on complex reasoning tasks. Extensive ablation studies reveal that these gains stem primarily from the recurrent inductive bias and strong nonlinear components of Transformer, rather than from overly complex architectural designs. Motivated by this insight, we propose the Universal Reasoning Model (URM), which enhances nonlinear depth-wise computation via short convolutional gating and improves optimization stability through truncated backpropagation through loops. URM achieves state-of-the-art performance on ARC-AGI 1 and 2.

\section{Acknowledgement}
We thank Benhao Huang for pointing out the typo in the previous version, and we also thank Zhengmao Ye from the Ubiquant AI team for providing infrastructure support.

\bibliography{references}{}
\bibliographystyle{plain}

\renewcommand{\thesubsection}{\Alph{subsection}}
\end{document}